\documentclass[10pt,twocolumn,letterpaper]{article}

\usepackage[preprint]{iccv}      
\usepackage{algorithm}
\usepackage{algpseudocode}
\usepackage{multirow} 
\usepackage{amsmath}
\definecolor{iccvblue}{rgb}{0.21,0.49,0.74}
\usepackage[pagebackref,breaklinks,colorlinks,allcolors=iccvblue]{hyperref}

\def\paperID{11340} 
\def\confName{ICCV}
\def\confYear{2025}

\title{Robust3D-CIL: Robust Class-Incremental Learning for 3D Perception}

\author{
Jinge Ma\\
Purdue University\\
\texttt{\small ma859@purdue.edu}
\and
Jiangpeng He\\
Massachusetts Institute of Technology\\
\texttt{\small jpenghe@mit.edu}
\and
Fengqing Zhu\\
Purdue University\\
\texttt{\small zhu0@purdue.edu}
}

\begin{document}
\maketitle
\begin{abstract}

3D perception plays a crucial role in real-world applications such as autonomous driving, robotics, and AR/VR. In practical scenarios, 3D perception models must continuously adapt to new data and emerging object categories, but retraining from scratch is impractical due to computational costs and time constraints. Therefore, adopting class-incremental learning (CIL) becomes particularly essential. In addition, real-world 3D point cloud data often include corrupted samples, which poses significant challenges for existing CIL methods and leads to more severe forgetting on corrupted data. To address these challenges, we consider the scenario in which a CIL model can be updated using point clouds with unknown corruption to better simulate real-world conditions. Inspired by Farthest Point Sampling, we propose a novel exemplar selection strategy that effectively preserves intra-class diversity when selecting replay exemplars, mitigating forgetting induced by data corruption. Furthermore, we introduce a point cloud downsampling-based replay method to utilize the limited replay buffer memory more efficiently, thereby further enhancing the model's continual learning ability. Extensive experiments demonstrate that our method improves the performance of replay-based CIL baselines by 2\% to 11\%, proving its effectiveness and promising potential for real-world 3D applications.

\end{abstract}    
\section{Introduction}
\label{sec:intro}

3D point cloud classification has a wide range of applications in real-world scenarios. For example, autonomous vehicles need to identify and classify surrounding objects in real-time~\cite{chen2019cooper}, while robots depend on 3D perception to understand and interact with their environment~\cite{jun2021towards}. However, real-world data always arrive sequentially over time with unseen objects and 3D point clouds are often corrupted due to sensor inaccuracies~\cite{10286105,pmlr-v162-ren22c,DBLP:journals/corr/abs-2201-12296}, highlighting the importance of both adaptivity and robustness in 3D perception models. Our motivation is to deploy 3D point cloud classification in such real-world environments, developing models that can continuously learn from both clean and corrupted 3D data stream while retaining previously learned knowledge. 

In real-world settings, point cloud data are significantly more prone to corruption than 2D images, largely due to the inherent limitations of 3D sensors and the complex nature of actual 3D environments~\cite{wu2019squeezesegv2,yan2020pointasnl}. Many studies have enhanced the robustness of classification models from the perspectives of model architecture~\cite{xiao2021triangle, yan2020pointasnl}, training methods~\cite{liu2021pointguard, dong2020self, sun2021adversarially}, and data augmentation~\cite{taghanaki2020robustpointset}. Class-incremental learning is a novel setup in which models need to learn from a continuous stream of data. However, a major challenge in CIL is catastrophe forgetting~\cite{MCCLOSKEY1989109}, where models tend to forget previously learned classes when trained on new tasks. To address this, existing CIL method~\cite{rebuffi2017icarl} usually employs a buffer to store a small number of learned task data as exemplars for knowledge replay, thus effectively mitigating the catastrophic forgetting issue.

\begin{figure}[h!]
    \centering
    \includegraphics[width=0.50\textwidth, height=0.26\textheight]{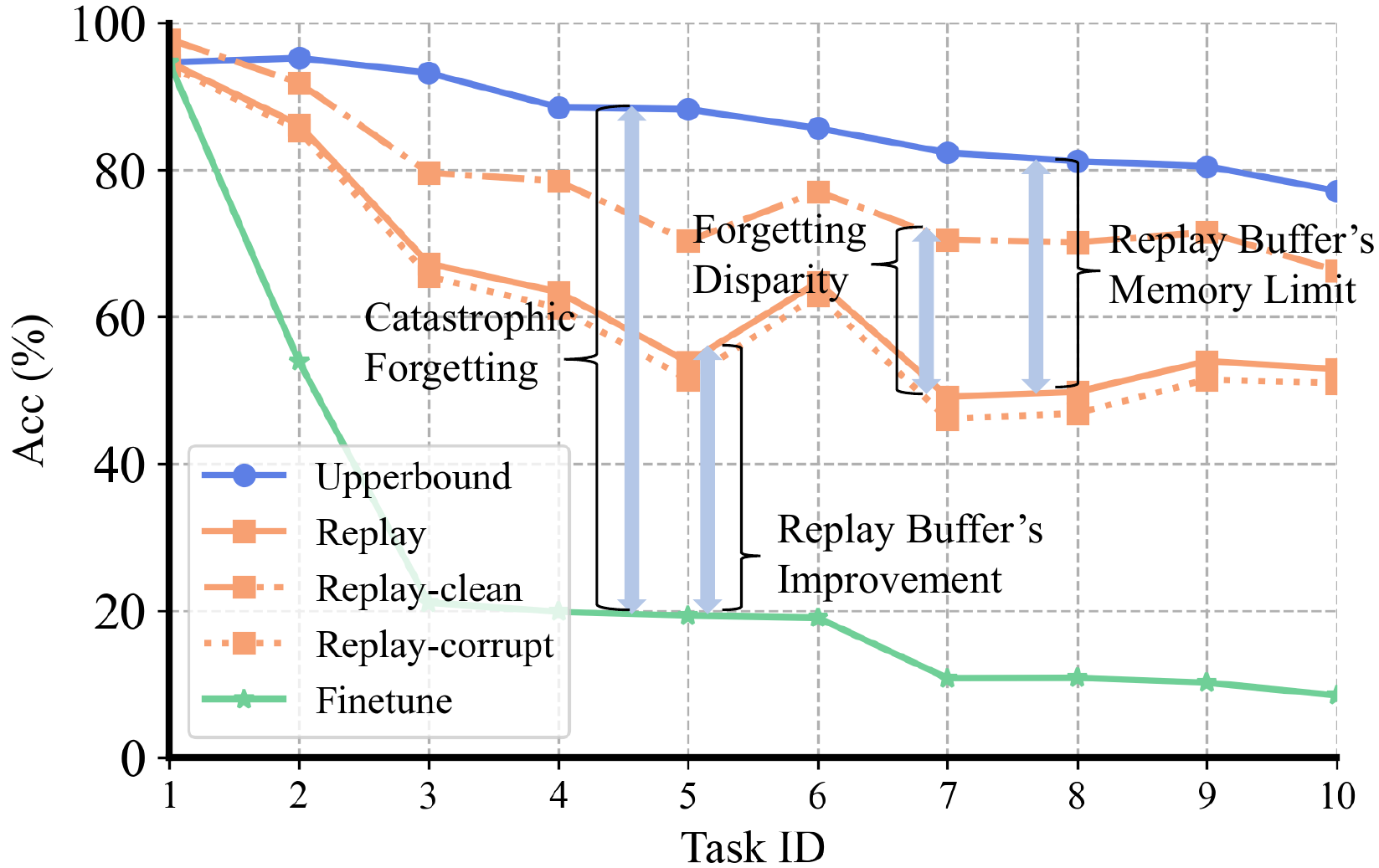}
    \caption{CIL performance of PointNeXt~\cite{qian2022pointnext} with partially corrupted ModelNet40~\cite{wu20153d}. The orange solid line (replay) in the figure is a weighted average of the two orange dashed lines (replay-clean and replay-corrupt). For details on the experimental setup, please refer to Sec.~\ref{scenario}}
    \vspace{-2mm}
    \label{fig1}
\end{figure}
Although there have been many studies in these two areas separately, CIL in the context of point cloud corruption remains under-explored. Our preliminary experiments have revealed unique challenges in this combined setting, as shown in~\cref{fig1}. We observe that even advanced 3D models exhibit catastrophic forgetting on simple datasets. Although the replay buffer significantly alleviates model's forgetting, the forgetting on corrupted point clouds is still more severe than on clean point clouds. Furthermore, the memory budget of the replay buffer limits its effectiveness: if an ideal replay buffer could store all previous training data, its performance would reach the upper bound.
We have shown the two main challenges in performing CIL methods for robust 3D perception:
\begin{enumerate}[label=(\arabic*)]
    \item Existing exemplar selection methods do not capture intra-class diversity of 3D point clouds, which may reduce model robustness on corrupted point clouds.
    \item Limited memory capacity for replay buffer restricts the number of exemplars that could be stored.
\end{enumerate}
To address these challenges, in this paper, we propose a standardized training/testing setup called Robust3D-CIL. 
Furthermore, we design an exemplar selection strategy inspired by the commonly used farthest point sampling algorithm~\cite{eldar1997farthest} for point cloud downsampling, which better covers the overall intra-class feature distribution and reflects intra-class differences. To better utilize the limited memory of the replay buffer, we exploit the unique characteristics of point cloud data. Specifically, point clouds are inherently unstructured and often contain redundant points, meaning that many points can be removed through downsampling without significantly compromising the overall geometric structure. By reducing the number of points per sample, we can store a larger number of training examples in the replay buffer, thereby improving the effectiveness of exemplar replay. Through extensive evaluations on the ModelNet40~\cite{wu20153d}, OmniObject3D~\cite{wu2023omniobject3dlargevocabulary3dobject}, and Objaverse-LVIS~\cite{deitke2023objaverse} datasets, we demonstrate that our approach outperforms existing methods in the context of Robust3D-CIL. Moreover, it can be integrated with various 3D feature extractors and CIL methods that employ replay buffers. 

In summary, our contributions are as follows:
\begin{itemize}
\item We propose a robust 3D point cloud class incremental learning setting, Robust3D-CIL, which better simulates real-world 3D point cloud scenarios.
\item We design a non-learning exemplar selection method to preserve intra-class diversity of training data better.
\item We propose an exemplar replay method based on point cloud downsampling, exploring the trade-off between the quality and quantity of replayed point clouds.
\end{itemize}

\section{Related Works}
\label{sec:related}

\subsection{Class Incremental Learning}

Class incremental learning is a paradigm of continual learning that aims to build a holistic classifier for all previously seen classes without relying on the task index during the inference stage. However, a significant challenge catastrophic forgetting occurs, where the model suffers severe performance degradation on earlier learned classes as new classes overwrite the previously acquired knowledge~\cite{9915459,wang2024few}. Many strategies have been proposed to address catastrophic forgetting in traditional 2D CIL, which can be broadly categorized into two types. 

\textbf{Dynamic network-based} methods prevent knowledge forgetting by adding neurons~\cite{Ostapenko_2019_CVPR,yoon2018lifelonglearningdynamicallyexpandable,NEURIPS2018_cee63112,li2019learn} or duplicating the backbone network~\cite{Yan_2021_CVPR,10.1007/978-3-031-19806-9_23,zhou2023model603exemplarsmemoryefficient,Aljundi_2017_CVPR,pmlr-v80-schwarz18a,Liu_2021_CVPR}, thereby preventing overwriting of previous knowledge and enhancing its representational power.
\textbf{Replay-based} methods store a buffer of exemplars from previous tasks and reuse them when learning new tasks. Non-learning-based replay, such as herding exemplar selection is widely used in~\cite{ILIO, Castro_2018_ECCV,Rebuffi_2017_CVPR,Wu_2019_CVPR} to select exemplars according to the distance to the class mean. More informative instances, such as those near class feature distribution margins~\cite{chaudhry2018riemannian} or instances with large prediction diversity~\cite{bang2021rainbow, he2021online_food} are also used. 
Some works improve the efficiency of memory buffers~\cite{zhou2022model} by storing extracted features~\cite{iscen2020memory, He_2022_WACV} or compressed versions that consume less storage capacity~\cite{yang2024probing, zhao2021memory}. 

\subsection{Robust 3D Perception}
Point clouds are a fundamental representation of 3D data, and point cloud classification is a core task in 3D perception. A variety of deep learning methods for point cloud classification have been widely explored, including multilayer perceptron (MLP)-based models~\cite{qi2017pointnet,qi2017pointnet++,qian2022pointnext}, convolution-based models~\cite{atzmon2018point,liu2019relation,xu2021paconv,wang2022learning}, graph-based models~\cite{simonovsky2017dynamic,wang2019dynamic,mohammadi2021pointview,lu2020pointngcnn}, attention-based models~\cite{guo2021pct,berg2022points,yu2022point,he2022voxel} and more recent pre-trained multi-modal large models~\cite{xue2023ulip,xue2024ulip,liu2023openshape,zhou2023uni3d}. However, due to the complexity of real-world environments, point clouds are often subject to significant corruption caused by sensors, environmental factors, and other interferences~\cite{pmlr-v162-ren22c,charron2018noising}. 

To mitigate the impact of corrupted point clouds on classification performance, several methods have been developed to enhance the robustness of point cloud classifiers~\cite{xiao2021triangle,yan2020pointasnl,liu2021pointguard,dong2020self}. Some studies, such as~\cite{ren2022benchmarking}, proposed standardized approaches to simulate real-world corruptions and conducted robustness evaluations on various representative point cloud classification models.

Prior work assumed full access to complete and clean point cloud datasets, which limits the adaptability in the complex real world where corrupted training data come in sequence. In this work, we are the first to apply robust 3D perception to the class-incremental learning scenario and enable more adaptable and robust 3D point cloud classification in dynamic CIL settings.

\begin{figure*}[!ht]
    \centering
    \includegraphics[width=1.05\textwidth]{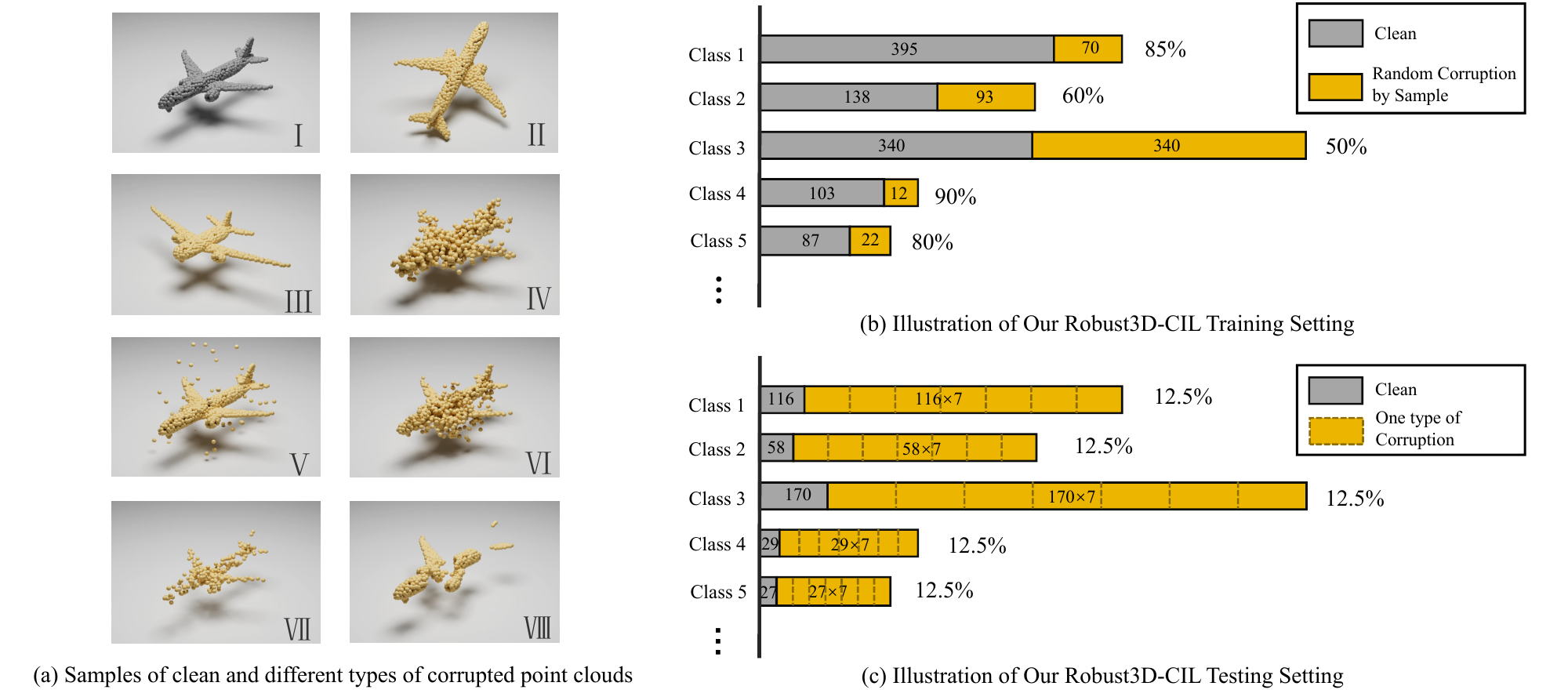}
    \caption{(a) Visualizations of clean point clouds and several common corrupted point cloud types are shown, specifically: \textbf{I} clean, \textbf{II} rotate, \textbf{III} scale, \textbf{IV} jitter, \textbf{V} add global, \textbf{VI} add local, \textbf{VII} dropout global, and \textbf{VIII} dropout local.(b) The training setup of \textbf{Robust3D-CIL} is displayed, where each class $C_1$ of the training point clouds includes a proportion, $\rho_k$, of clean point clouds ($\rho_k \in [0.5, 0.95]$), with the remaining being randomly corrupted point clouds of various types. The corruption of the training point clouds and $\rho_k$ is invisible to the model, simulating the real-world scenario where training point clouds may contain unknown corruptions. (c) The testing setup of \textbf{Robust3D-CIL} is illustrated, where each class includes both clean point clouds and an equal proportion of each type of corrupted point cloud, to better assess the model's robustness across various test data conditions.}
    \label{fig2}
\end{figure*}
\subsection{3D Class Incremental Learning}
Recent studies on CIL for 3D point clouds have gradually attracted attention, although the area remains relatively under-explored. Inor-net~\cite{dong2023inornet} leveraged 3D geometric information to learn distinctive features for each class and correct biased weights caused by class imbalance, thereby mitigating forgetting. Chowdhury et al.~\cite{chowdhury2022fewshot} focused on few-shot 3D CIL by constructing a set of basic descriptions to enhance the model’s adaptability. L3doc~\cite{liu2021l3doc} proposed a framework that enables the model to retain existing knowledge while continuously incorporating new classes. Tan et al.~\cite{tan2024crossdomain} addressed the challenge of domain shifts with a cross-domain few-shot incremental learning approach, and Zhao et al.~\cite{zhao2022static} introduced a static-dynamic co-teaching strategy to simultaneously preserve old knowledge and learn new information.

Existing work has not explicitly addressed the robustness challenges posed by real-world point cloud corruption. In our work, as mentioned in~\cref{sec:intro}, we improve the traditional replay buffer to account for the unique characteristics of 3D point cloud data, and our method can be flexibly integrated with existing CIL approaches to enhance overall robustness.

\begin{figure*}[ht]
    \centering
    \includegraphics[width=0.9\linewidth]{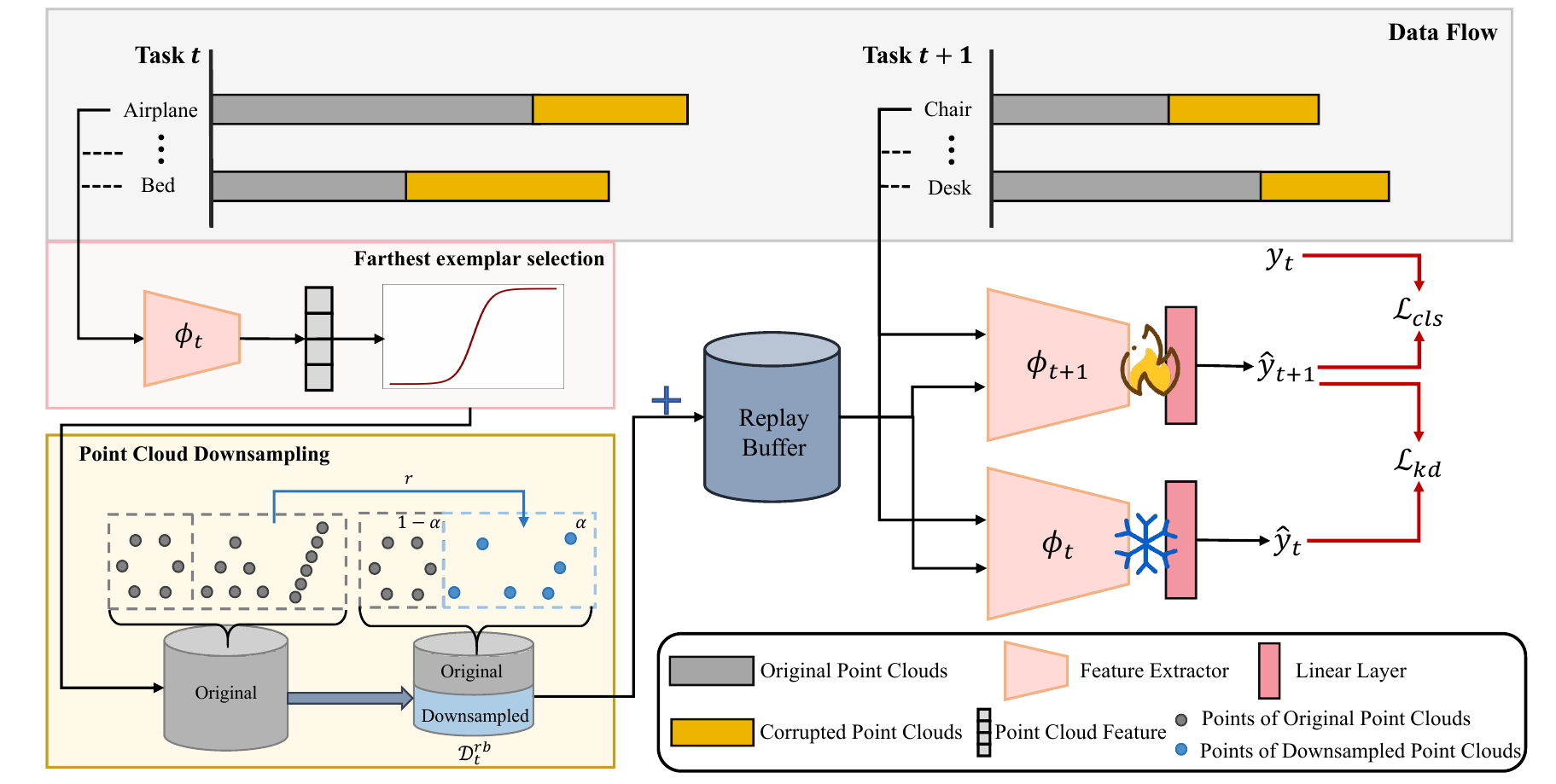}
    \caption{An overview of our method: after completing the learning of task \( t \), we use the current model to extract point cloud features and then apply \textbf{farthest exemplar selection} to choose replay exemplars that are both representative and show intra-class diversity. Then we \textbf{downsample} a proportion \( \alpha\) of the exemplars with downsampling rate \( r\) to reduce the required storage, to enable more replay exemplars to fit within the limited replay buffer memory. For most baseline models, in addition to the classification loss, other losses such as knowledge distillation loss are also included.}
    \label{fig3}
    \vspace{-5mm}
\end{figure*}
\section{Preliminaries}
\label{prelim}
In class incremental learning (CIL), a neural network incrementally learns from multiple tasks with additional classes, where each task dataset is denoted by \( \mathcal{D}_t =\bigcup_{k\in K_t}\{(P_{k,i}, y_{k,i})\}_{i=1}^{m_k} \). Here, \( \mathcal{D}_t \) is only available during training on task \( t \), where \( (P_{k,i}, y_{k,i}) \) represents a point cloud sample and its label. \(m_k\) denotes the number of training samples in $k$-th category, and \(K_t\) is the set consisting of the indices of newly introduced classes in task \( t \). For classification tasks, training samples within each task can contain one or more new classes. During testing, the model needs to evaluate all previously seen classes to distinguish categories from different tasks~\cite{li2017learning}.

Assume that a DNN with parameters \( \theta \) has already learned \( T \) tasks and is now trying to learn a new task \( T+1 \). Since the previous training data \( \bigcup_{t=1}^T \mathcal{D}_t \) are no longer available, the DNN can only adapt its learned parameters to fit the new dataset \( \mathcal{D}_{T+1} \), leading to catastrophic forgetting, where the model performs well on the new task but significantly degrades on previously learned tasks.

In the standard CIL setup, the class set \( \mathcal{C} = \{C_1, C_2, \dots, C_n\} \) is divided into multiple tasks, each containing a subset of categories \( \mathcal{T}_t \). During each task \( t \), the model learns samples from the new categories in \( \mathcal{T}_t \) while retaining knowledge of previously learned classes. After training on task \( t \), the model is evaluated on all classes learned up to that point, denoted as the test set \( \mathcal{D}_{\text{test}}^{(1:t)} \), to assess catastrophic forgetting and generalization.

We often observe that as the number of training tasks increases, the model's accuracy on learned classes gradually decreases. This occurs because the model forgets previous knowledge while learning new tasks. This is known as \textbf{catastrophic forgetting}. 

\section{Robust3D-CIL Scenario}
\label{scenario}
\subsection{Scenario Setting}
In the real world, point cloud data often suffer from various types of corruption, posing a significant challenge to model robustness. To better simulate this situation, as shown in~\cref{fig2}, we create hybrid datasets of clean and corrupted point clouds based on clean point cloud datasets. In our Robust3D-CIL scenario setting, the training and testing datasets are constructed as follows:

Let a category set be denoted by $\mathcal{C} = \{C_1, C_2, \dots, C_n\}$, where the $k$-th category has a point cloud dataset $\mathcal{D}_{C_k} = \{P_{k,1}, P_{k,2}, \dots, P_{k,m_k}\}$, with $m_k$ representing the number of point cloud samples in category $C_k$. 
\begin{itemize}
    \item For the training dataset, we sample an unfixed proportion $\rho_k$(randomly chosen between 0.5 and 0.95) of point clouds from each category to generate corrupted versions. The corruption types include seven methods: \textbf{scale} (denoted as $S$), \textbf{rotate} ($R$), \textbf{jitter} ($J$), \textbf{add global} ($A_G$), \textbf{add local}($A_L$), \textbf{ dropout global}($D_G$), and \textbf{dropout local}($D_L$). Each sampled point cloud is corrupted by only one random type, then mixed with the remaining clean point clouds to form the training dataset $\mathcal{D}_{\text{train}}$. To mimic real-world scenarios, the presence and type of corruption are not visible to the model during training.
    
    \item For the test dataset, we retain the original clean samples $\mathcal{D}_{C_k}^{\text{clean}}$ for each category and generate seven corresponding  corrupted versions for each point cloud $\{P_{k,i}^S, P_{k,i}^R, P_{k,i}^J, P_{k,i}^{A_G}, P_{k,i}^{A_L}, P_{k,i}^{D_G}, P_{k,i}^{D_L}\}$. Thus, the test dataset is 8 times the original size, denoted as $\mathcal{D}_{\text{test}} = \bigcup_{k=1}^n \bigcup_{i=1}^{m_k} \{P_{k,i}, P_{k,i}^S, P_{k,i}^R, P_{k,i}^J, P_{k,i}^{A_G}, P_{k,i}^{A_L}, P_{k,i}^{D_G}, P_{k,i}^{D_L}\}$. Each point cloud and its seven corrupted versions exist with a uniform distribution.
\end{itemize}

\subsection{Point Cloud Corruption}

Since it is challenging to control real-world point cloud corruption, we adopt the corruption methods proposed by~\cite{ren2022benchmarking}. In their work, corruption operations are strictly prohibited during training to meet OOD evaluation principles. However, our objective is not to evaluate existing 3D point cloud models' generalization to OOD data but to simulate a real-world mixture of clean and corrupted point clouds. Therefore, our training dataset includes both clean and corrupted point clouds. Regarding corruption levels, \cite{ren2022benchmarking} introduced five severity levels, resulting in a test set 35 times larger (5 levels, 7 corruption types), which would significantly increase the testing cost for each task in a CIL scenario. To save computational resources and ensure reasonable settings, we only use the highest level of corruption to test model performance under strict conditions more effectively. For more details, please refer to the Sec. A and B in the appendix.

\section{Methodology}
Our method combines example replay based on point cloud downsampling and farthest exemplar selection, aiming to optimize memory usage and enhance the representativeness of stored samples. \cref{fig3} shows the overall process of our method. As mentioned in~\cref{sec:intro}, to address two major challenges in Robust3D-CIL to achieve robust CIL, we proposed two methods accordingly:

\noindent \textbf{Representative Replay Exemplar Selection:}
We propose a farthest exemplar selection strategy to balance the probability of selecting under-represented samples.

\noindent\textbf{Limited Replay Buffer Memory:}
We propose a replay strategy based on point cloud downsampling that stores more exemplars while keeping memory budget unchanged. Additionally, we explore the trade-off between the quantity and quality of stored point cloud exemplars.

\subsection{Farthest Exemplar Selection}
\begin{figure}[ht]
    \centering
    \includegraphics[width=0.4\textwidth, height=0.34\textheight]{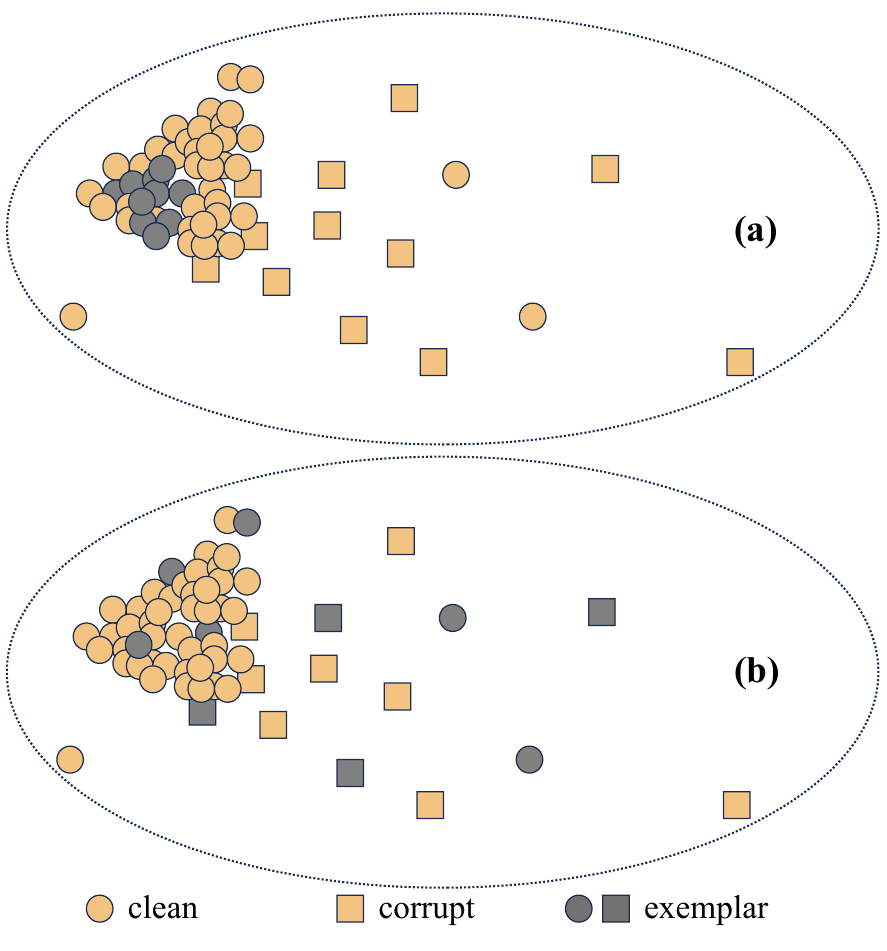}
    \caption{This is a 2D visualization of the feature distribution for a particular class. (a) shows \textbf{herding selection}, in which at each step the sample closest to the center of the remaining unselected samples is chosen as the exemplar, resulting in a predominance of clean samples.  (b) shows our proposed \textbf{farthest selection}, where each selected exemplar suppresses the likelihood of nearby samples being chosen, driving the exemplars to better cover the entire distribution and capture the intra-class variance.}
    \label{fig4}
\end{figure}
In class-incremental learning, a common approach is to use a replay buffer \( \mathcal{D}_{t}^{\text{rb}} = \{(P_{t,i}, y_{t,i})\}_{i=1}^{n_t^{\text{rb}}} \) to store representative samples from previous tasks and replay these samples during new task training. 

To select representative samples from the training data, we first extract features of each sample through a feature extraction network \( F_\theta^e(\cdot) \). In herding exemplar selection, exemplars are chosen based on their proximity to the mean of the currently unselected samples \( \mu_C \) in the feature space. ~\cref{fig4} shows a schematic illustration of a intra-class point cloud feature distribution. We observed that the feature distribution of clean point clouds is more concentrated, whereas that of corrupted point clouds is more dispersed. \textbf{Most CIL methods} employ herding selection, which tends to select exemplars biased towards clean point clouds, thus failing to adequately capture the intra-class variance.

To address this issue, we propose the \textbf{farthest exemplar selection} strategy. Specifically, the steps of the farthest exemplar selection are as~\cref{fes}:

\begin{algorithm}
\caption{Farthest Exemplar Selection}
\label{fes}
\begin{algorithmic}[1]
\Require Class feature set \( \mathcal{F}_C = \{F_\theta^e(P) \mid P \in \mathcal{D}_{C}\} \), number of exemplars to select \( N \)
\Ensure Selected exemplar set \( S_C \) with \( |S_C| = N \)

\State Initialize selected set \( S_C = \{\} \)
\State Randomly select an initial sample \( P_0 \) from \( \mathcal{F}_C \)
\State Add \( P_0 \) to \( S_C \)

\While{\( |S_C| < N \)}
    \ForAll{\( P_i \in \mathcal{F}_C \setminus S_C \)}
        \State \( d_i = \min_{P_j \in S_C} \|F_\theta^e(P_i) - F_\theta^e(P_j)\| \)
    \EndFor

    \State \( \mathbf{d} \leftarrow \mathbf{d} / \max(\mathbf{d}) \)
    \State \( \mathbf{p} = f(\mathbf{d}) \)
    \State \( \mathbf{p} \leftarrow \mathbf{p} / \sum \mathbf{p} \)
    \State Sample \( P_{\text{next}} \) from \( \mathcal{F}_C \setminus S_C \) using \( \mathbf{p} \)
    \State \( S_C \leftarrow S_C \cup \{P_{\text{next}}\} \)
\EndWhile

\Return \( S_C \)
\end{algorithmic}
\end{algorithm}

Note: The vector \( \mathbf{d} \) is composed of elements \( d_i \), where \( d_i \) represents the minimum distance between an unselected sample \( P_i \) and the \( S_C \). The distance-based probability function \( f(d) \) such that the probability \( p_i = f(d_i) \) of selecting a sample should be a positive function. Different choices of \( f(d) \) lead to different selection behaviors:
\begin{itemize}
    \item  When \( f(d_i) \) is an indicator function, farthest exemplar selection reduces to \textbf{farthest point sampling}:
\begin{equation}
   f(d_i) = \mathbf{1}_{\{d_i = \max(d)\}} = 
     \begin{cases} 
     1, & \text{if } d_i = \max(d) \\
     0, & \text{otherwise}
     \end{cases} 
\end{equation}

    \item When \( f(d_i) \) is a constant, farthest exemplar selection reduces to \textbf{random sampling}:
\begin{equation}
    f(d_i) \equiv c, \quad \forall d_i
\end{equation}

\end{itemize}
We finally define \( f(d) \) as a monotonically increasing function, specifically the Sigmoid function. This encourages samples that are farther from the currently selected set to have a higher probability of being chosen, effectively suppressing the likelihood of selecting samples that are close to those already in the set. 

To introduce randomness in the selection process, we further control the randomness with a constant \( c \). Therefore, we define \( f(d) \) as:
\begin{equation}
    f(d_i) = \text{Sigmoid}(d_i; k, c) = \frac{1}{1 + \exp(-k (d_i - \bar{d}))} + c
\label{eq:fes}
\end{equation} 
where \( k \) controls the steepness of the Sigmoid function, \( \bar{d} \) is the mean of normalized distances.

By using this selection method, as shown in ~\cref{fig4}, farthest exemplar selection effectively covers the in-class distribution in feature space, ensuring intra-class diversity in the replay buffer and thus improving model robustness.  For the actual visualizations of more exemplar selection strategies, please refer to Appendix Fig. 7

\subsection{Replay with Point Cloud Downsampling}

To address limited replay buffer memory, we propose incorporating a \textbf{point cloud downsampling} strategy in the exemplar replay buffer. 

Given a point cloud dataset for class \( C \), denoted by \( \mathcal{D}_C = \{P_{t,i}\}_{i=1}^{n_t^{\text{rb}}} \), where each point cloud \( P_{t,i} \) consists of multiple points, our replay buffer consists of two versions of each point cloud: the \textbf{original point cloud} and the \textbf{downsampled point cloud}. The downsampled point cloud, denoted by \( P_{t,i}^{\text{down}} \), is obtained using farthest point sampling. The replay buffer can be defined as:
\begin{equation}
   \mathcal{D}_{t}^{\text{rb}} = \{(P_{t,i}, P_{t,i}^{\text{down}}, y_{t,i})\}_{i=1}^{n_t^{\text{rb}}}
\end{equation}
where \( P_{t,i} \) represents the original point cloud, \( P_{t,i}^{\text{down}} \) is the downsampled point cloud, and \( y_{t,i} \) is the corresponding class label.

In this design, we introduce two key variables to control point cloud downsampling in the replay buffer: \textbf{Downsampled Point Cloud Ratio} \( \alpha \) and \textbf{Downsampling Rate} \( r \). \( \alpha \) represents the fraction of downsampled point clouds within the replay buffer for each class, and \( r \) is defined as the ratio of the number of points in the original point cloud to that in the downsampled point cloud: 
\begin{equation}
    \alpha = \frac{|\{P_{t,i}^{\text{down}}\}_{i=1}^{n_t^{\text{rb}}}|}{|\mathcal{D}_{t}^{\text{rb}}|}, \quad
r = \frac{|P_{t,i}|}{|P_{t,i}^{\text{down}}|}
\label{eq:alpha_ratio}
\end{equation}
where \( |\{P_{t,i}^{\text{down}}\}_{i=1}^{n_t^{\text{rb}}}| \) denotes the number of downsampled point clouds, \( |\mathcal{D}_{t}^{\text{rb}}| \) represents the total number of samples in the replay buffer, \( |P_{t,i}^{\text{down}}| \) is the number of points in the downsampled cloud, and \( |P_{t,i}| \) is the number of points in the original cloud.

Given the total number of points \( M \) that can be stored in the buffer for each class, the number of exemplars \( N \) per class that can be stored in the buffer can be formulated as:

\begin{equation}
    N = \frac{Mr}{n \cdot ((1 - \alpha)r + \alpha)}
\label{eq:alpha_ratio2}
\end{equation}
where \(n = 1024\) represents the number of points in each original point cloud.

Increasing \( \alpha \) or \( r \) allows storing more samples within limited memory, although it reduces the fidelity of the original point cloud. We would further analyze the quality-quantity trade-off in~\cref{experiments}.

\section{Experiments}
\label{experiments}
In this section, we introduce the experimental setup, including the training and testing datasets, backbone network, and hyperparameters. We also analyze the performance of our method compared to baseline models. For the ablation study on farthest exemplar selection and point cloud downsampling, please refer to the supplementary material.

\begin{figure*}[ht]
    \centering
    \includegraphics[width=0.9\textwidth, height=0.3\textheight]{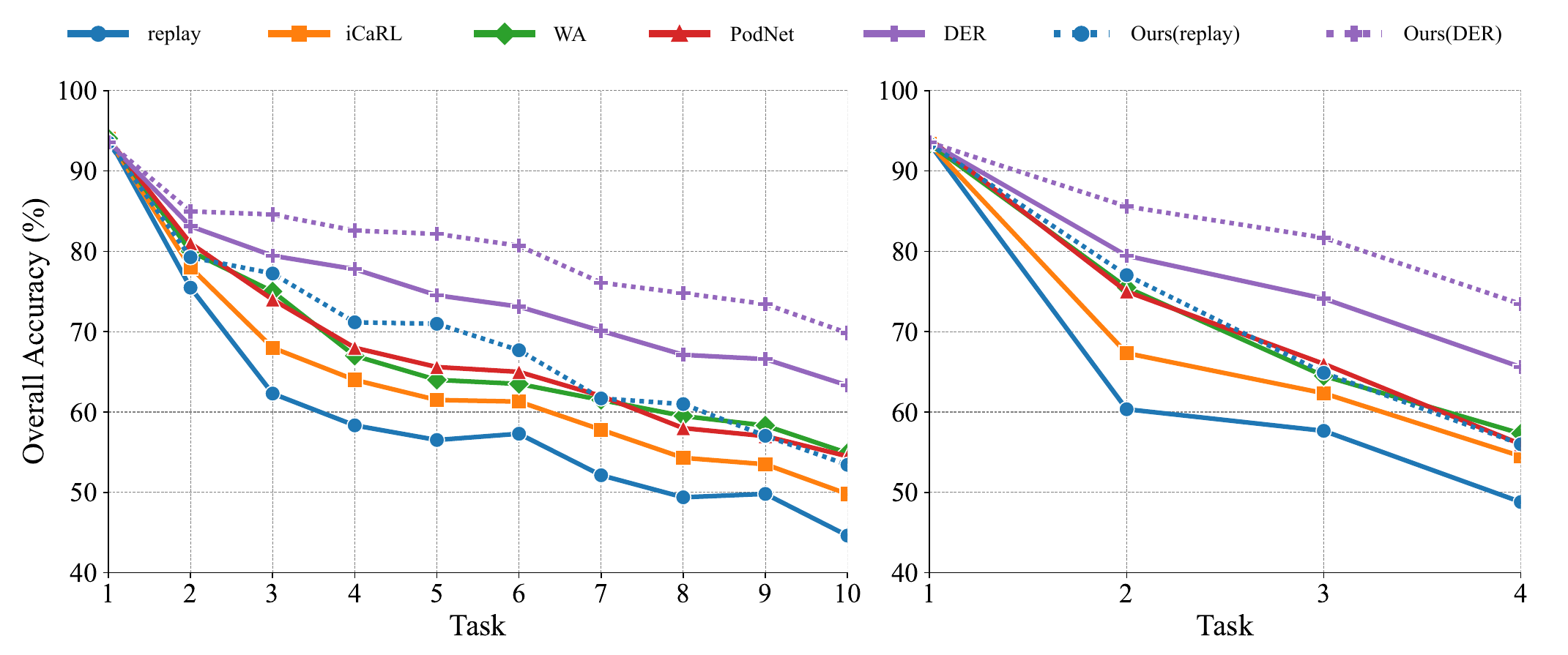}
    \caption{These plots show the performance of the baseline models and our method on ModelNet40-Mix. The left plot has 10 tasks, and the right plot has 4 tasks. Our method significantly improves the robustness of the baseline models.}
    \label{fig5}
    \vspace{-3mm}
\end{figure*}

\subsection{Datasets}

We generated our training/testing datasets using the ModelNet40~\cite{wu20153d} and OmniObject3D~\cite{wu2023omniobject3dlargevocabulary3dobject} and Objaverse-LVIS datasets. ModelNet40 is a widely used CAD dataset in 3D perception research, comprising 40 object categories with a total of 9,840 training point clouds and 2,468 testing point clouds; OmniObject3D is a real-scanned 3D dataset covering 216 categories with a total of 5,911 point clouds; and Objaverse-LVIS is a large labeled dataset extracted from the Objaverse dataset, containing 1,156 categories and 46,054 point clouds. For both OmniObject3D and Objaverse-LVIS, we adopt the same processing strategy: for each category, we randomly select 80\% of the point clouds as training data, with the remaining 20\% reserved for testing. All these original point clouds are considered high-quality and clean. Based on the process described in Robust3D-CIL, we introduce random corruptions to some of the clean point clouds to construct the ModelNet40-Mix, OmniObject3D-Mix, and Objaverse-LVIS-Mix datasets for actual training and testing. For more details on the types and levels of point cloud corruption, please refer to Sec.~\ref{scenario} and the appendix.

\subsection{Experimental Setup}

We mainly used PointNet++~\cite{qi2017pointnet++} as the feature extraction network for the main experiments. Additionally, we conducted experiments using DGCNN~\cite{wang2019dynamic} and PointNeXt~\cite{qian2022pointnext} as backbones to enhance the credibility of our method. We removed the built-in data augmentation in its original implementation. The number of tasks was set to 4, 10, and 20 for both datasets. The batch size was set to 64, and the test batch size was set to 128. Each task was trained for 50 epochs. For the parameters in farthest exemplar sampling, we set the Sigmoid steepness control variable \( k = 20 \) and randomness control variable \( c = 0.2 \) in eq. \ref{eq:fes}. In replay with point cloud downsampling, as shown in eq. \ref{eq:alpha_ratio} and \ref{eq:alpha_ratio2}, we set \( \alpha = 0.5 \), \( r = 2 \), with \( M = 20 \times 1024 \), \( N = 30 \) for ModelNet40-Mix , \( M = 5 \times 1024 \), \( N = 10 \) for OmniObject3D-Mix, and \( M = 10 \times 1024 \), \( N = 15 \) for Objaverse-LVIS-Mix. We used different replay memory per class for these datasets because the number of training samples per class varies significantly between them. 
Following the evaluation convention of point cloud classification tasks, we selected Overall Accuracy (OA) as the standard for evaluating model performance.

\subsection{Baselines and Our Methods}

\begin{table*}[!ht]
\centering
\scalebox{0.9}{
\begin{tabular}{lccccccccc}
\toprule
\multirow{2}{*}{Method} & \multicolumn{3}{c}{ModelNet40-Mix} & \multicolumn{3}{c}{OmniObject3D-Mix} & \multicolumn{3}{c}{Objaverse-LVIS-Mix} \\ \cmidrule(lr){2-4} \cmidrule(lr){5-7} \cmidrule(lr){8-10}
                      & 4-task & 10-task & 20-task & 4-task & 10-task & 20-task & 4-task & 10-task & 20-task \\ \midrule
Replay      & 65.05 & 59.99 & 56.63 & 44.68 & 42.79 & 41.97 & 23.34 & 21.53 & 20.92 \\
iCaRL       & 69.39 & 64.23 & 62.05 & 48.59 & 47.83 & 46.77 & 25.18 & 23.96 & 23.10 \\
PODNet      & 72.67 & 67.78 & 63.80 & 50.69 & 49.92 & 49.15 & 25.71 & 24.23 & 23.89 \\
WA          & 72.74 & 67.91 & 63.55 & 50.73 & 50.68 & 49.63 & 26.15 & 24.13 & 23.91 \\
DER         & 78.19 & 74.87 & 70.07 & 53.38 & 52.57 & 52.14 & 27.88 & 27.45 & 27.17 \\ \cmidrule(lr){1-10}
Ours+Replay & 72.82{\fontsize{6}{10}\selectfont (+7.77)} & 69.31{\fontsize{6}{10}\selectfont (+9.32)} & 67.78{\fontsize{6}{10}\selectfont (+11.15)} & 49.59{\fontsize{6}{10}\selectfont (+4.91)} & 48.18{\fontsize{6}{10}\selectfont (+5.39)} & 48.03{\fontsize{6}{10}\selectfont (+6.06)} & 25.76{\fontsize{6}{10}\selectfont (+2.42)} & 24.52{\fontsize{6}{10}\selectfont (+2.99)} & 24.18{\fontsize{6}{10}\selectfont (+3.26)}\\
Ours+DER    & \textbf{83.57}{\fontsize{6}{10}\selectfont (+5.38)} & \textbf{81.27}{\fontsize{6}{10}\selectfont (+6.40)} & \textbf{78.13}{\fontsize{6}{10}\selectfont (+8.06)} & \textbf{57.05}{\fontsize{6}{10}\selectfont (+3.67)} & \textbf{56.75}{\fontsize{6}{10}\selectfont (+4.18)} & \textbf{57.15}{\fontsize{6}{10}\selectfont (+5.01)} & \textbf{29.85}{\fontsize{6}{10}\selectfont (+1.97)} & \textbf{29.60}{\fontsize{6}{10}\selectfont (+2.15)} & \textbf{29.53}{\fontsize{6}{10}\selectfont (+2.36)} \\ \bottomrule
\end{tabular}
}
\caption{Performance comparison of different baselines and our method on the ModelNet40-Mix, OmniObject3D-Mix, and Objaverse-LVIS-Mix datasets. Our method consistently achieves the best performance across all datasets and task settings, with greater improvements observed as the number of tasks increases. The metric in the table represents the average overall accuracy (\%) across all tasks.}
\label{tab1}
\vspace{-3mm}
\end{table*}

We evaluated several representative baseline models in the Robust3D-CIL setting, including exemplar replay-based methods such as iCaRL\cite{rebuffi2017icarl}, PODNet\cite{douillard2020podnet}, WA\cite{zhao2020maintaining}, and DER\cite{yan2021dynamically} and the replay-only baseline. All methods were implemented using the PyCIL\cite{zhou2023pycil} library, and the default hyperparameters were used to ensure a fair comparison.

Our method could be flexibly combined with existing CIL methods. To demonstrate the effectiveness of our approach, we combine farthest exemplar selection and replay with point cloud downsampling with both the replay-only method and DER, and evaluate the resulting improvements.

\subsection{Results Analysis}
Figure~\ref{fig4} and Table~\ref{tab1} present the comparison between the baseline models based on exemplar replay and our proposed method. The experiments demonstrate that our method significantly improves the performance of replay-only methods and DER under the Robust3D-CIL setting. On ModelNet40-Mix, our method achieves an average improvement of 9.41\% over the replay-only method and 6.61\% over DER. On OmniObject3D-Mix, our method achieves an average improvement of 5.45\% over the replay-only method and 3.95\% over DER. This demonstrates the effectiveness of our approach. We compared the improvements across different numbers of tasks and found that the more tasks there are, the greater the improvement achieved by our method. We analyze that this is because, for replay-based methods, the larger the number of tasks, the more frequently replay exemplars are retrained. The model’s performance becomes more affected by the quality of the replay exemplars. Therefore, an exemplar selection method that can reasonably represent intra-class diversity contributes more to enhancing model robustness.

\subsection{Different Backbones}
\begin{table}[!ht]
\centering
\scalebox{0.9}{
\begin{tabular}{lcccc}
\toprule
\multirow{2}{*}{Method} & \multicolumn{2}{c}{DGCNN} & \multicolumn{2}{c}{PointNeXt} \\ \cmidrule(lr){2-3} \cmidrule(lr){4-5}
                        & 4-task & 20-task & 4-task & 20-task \\ \midrule
Replay      & 64.97 & 56.26 & 64.84 & 56.44 \\
iCaRL       & 69.34 & 61.97 & 69.09 & 61.83 \\
PODNet      & 72.83 & 64.00 & 72.70 & 64.19 \\
WA          & 72.62 & 63.86 & 72.82 & 63.18 \\
DER         & 77.95 & 69.82 & 78.13 & 69.95 \\ \midrule
Ours+DER    & \textbf{83.21}{\fontsize{6}{10}\selectfont (+5.26)} & \textbf{77.72}{\fontsize{6}{10}\selectfont (+7.90)} & \textbf{82.93}{\fontsize{6}{10}\selectfont (+4.80)} & \textbf{77.34}{\fontsize{6}{10}\selectfont (+7.39)} \\ \bottomrule
\end{tabular}
}
\caption{The performance differences among different backbones on ModelNet40-Mix are minimal, indicating the backbone structure has little impact on its resistance to forgetting under CIL. However, we observed that different backbones require a different number of epochs to converge. The results show the effectiveness of our method does not depend on a specific backbone.}
\label{tab2}
\vspace{-4mm}
\end{table}

Table~\ref{tab2} demonstrates the influence of different backbone networks on model performance. We keep all other experimental settings unchanged and conduct experiments using three different backbones, including PointNet++, DGCNN, and PointNeXt. For a fair comparison under Robust3D-CIL, we removed all data augmentation during the backbone input stage. The results indicate that even advanced backbones suffer from catastrophic forgetting. Moreover, the effectiveness of our method is not backbone-specific but rather data-centric.

\subsection{Quantity-Quality Trade-off}

We further analyzed the trade-off between exemplar quantity and quality, as shown in Figure \ref{fig5}. Using the basic replay method, we examined the effect of parameters $\alpha$ and $r$ on performance, where $\alpha$ represents the exemplar selection parameter and $r$ are defined in eq. \ref{eq:alpha_ratio}. We plotted the OA-$\alpha$ curves for $r = 2, 4, 8, 16$. The results show that an appropriate combination of $r$ and $\alpha$ can significantly outperform the original replay buffer, which are the far left points of the the OA-$\alpha$ curves. For $\alpha$, we observe a peak in the curves. Both excessively high and low values of $\alpha$ negatively impact performance. For different $r$ values, the corresponding optimal $\alpha_0$ varies. We observe that larger $r$ values generally correspond to smaller $\alpha_0$, possibly because the benefit from quantity is more limited by the declined fidelity of the severely downsampled point clouds. As $r$ decreases, the OA-$\alpha$ curve becomes flatter, as the downsampled point clouds are more similar to the original point clouds, making the effect of $\alpha$ less significant.

 These experimental results demonstrate how to empirically adjust $\alpha$ and $r$ to balance the quantity and quality of replay exemplars within limited memory, thereby maximizing the effectiveness of the replay buffer's memory.

\begin{figure}[h!]
    \centering
    \includegraphics[width=0.49\textwidth, height=0.32\textheight]{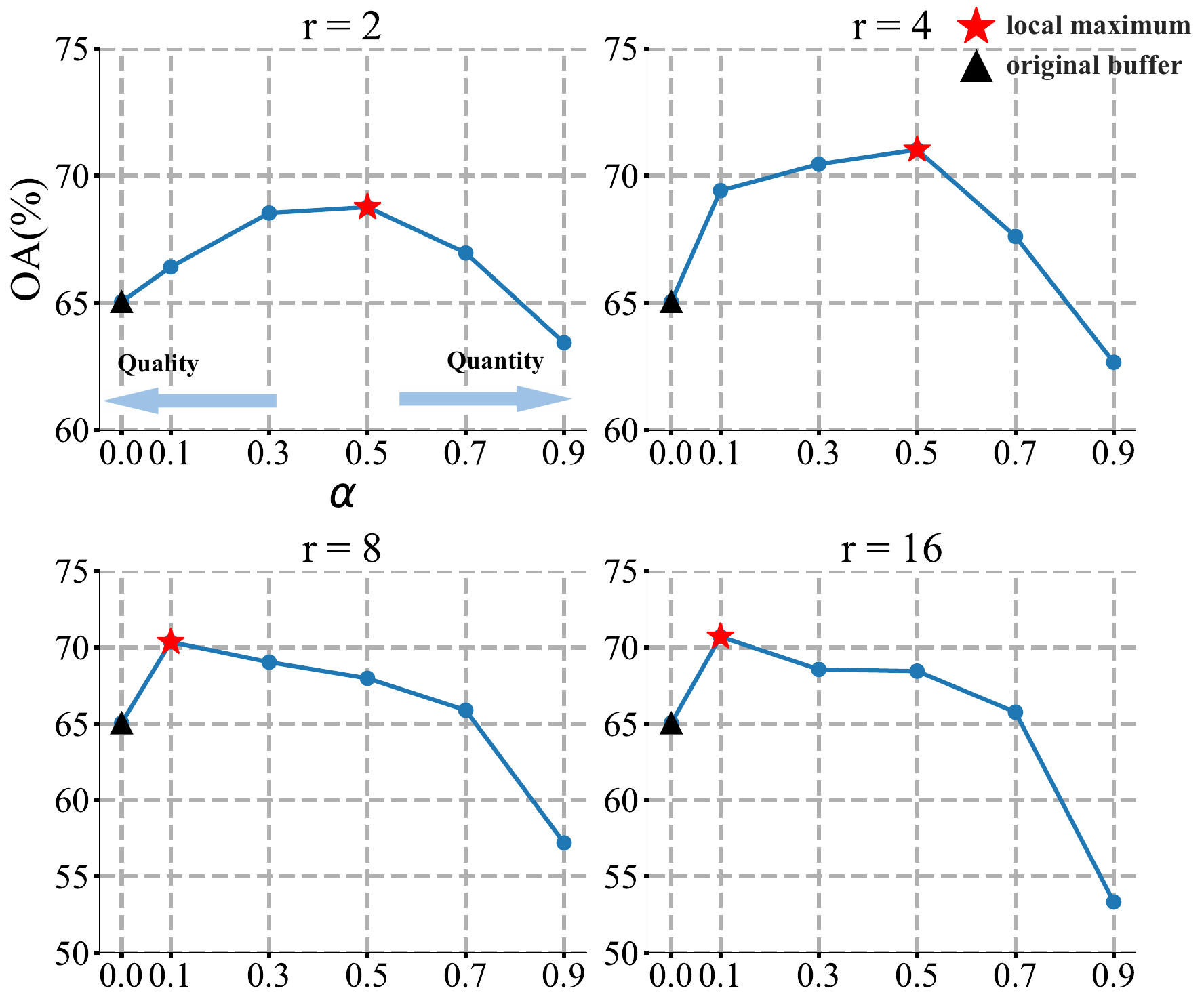}
    \caption{Quality-quantity trade-off: Experiments were conducted on ModelNet40-Mix with a task number of 10, using our farthest exemplar selection. The leftmost point of the curve represents the original replay buffer without downsampling. As the downsampled point cloud ratio $\alpha$ increases, the number of replay exemplars grows, but the average quality decreases.(Please zoom in)}
    \vspace{-3mm}
    \label{fig6}
\end{figure}

\section{Conclusions}
In this study, we propose a novel robust class incremental learning setting for 3D point cloud classification—Robust3D-CIL. Robust3D-CIL aims to simulate the robustness challenges of applying 3D perception models in real-world point cloud data streams. To address these challenges, we leverage point cloud downsampling to improve memory efficiency in the replay buffer, exploring the balance between the quality and quantity of replay exemplars. Additionally, we design a new exemplar selection strategy based on the point cloud downsampling algorithm (farthest exemplar selection), allowing replay exemplars to more effectively capture intra-class diversity. Our method enhances the robustness of replay-based methods, mitigates the impact of catastrophic forgetting, and adapts to mixed clean and corrupted data. We hope this study will inspire further research on robust class incremental learning methods in 3D perception and other downstream tasks.

{
    \small
    \bibliographystyle{ieeenat_fullname}
    \bibliography{main}
}
\clearpage
\setcounter{page}{1}
\maketitlesupplementary
\appendix
\section{Point Cloud Post-processing}

Among the seven types of corruption, scaling ($S$), rotation ($R$), and jittering ($J$) maintain the same number of points, while global addition ($A_G$), local addition ($A_L$), global dropout ($D_G$), and local dropout ($D_L$) alter the point count. To enable batch processing for 3D perception models and to simulate real-world cases where corruption cannot be identified by the number of points alone, we preprocess all corrupted point clouds to unify their point counts.

Specifically, we set the number of points for all point clouds to 1024: for corrupted point clouds with more than 1024 points, we use farthest point sampling~\cite{eldar1997farthest} to select 1024 points; for point clouds with fewer than 1024 points, we repeat the first point until reaching 1024 points. This approach enables batched training while keeping the original geometry information of the corrupted point clouds.

\section{Simulation of Point Cloud Corruptions}

We utilized the methods and code from this paper \cite{pmlr-v162-ren22c} to simulate real-world point cloud corruptions, ensuring that all settings were consistent with the original work. However, applying all combinations of corruption types and severity levels would increase the size of the test dataset 35-fold, as each type of 7 corruptions with its 5 severity levels doubles the data size. To reduce testing costs, we retained only the highest severity level for each corruption type, as it effectively represents the characteristics of corrupted point clouds.

\paragraph{Jitter}
Gaussian noise $\epsilon \sim \mathcal{N}(0, \sigma^2)$ is added to each point's X, Y, and Z coordinates, where $\sigma \in \{0.01, 0.02, 0.03, 0.04, 0.05\}$ represents five severity levels. For our experiments, we retained only the most severe level, $\sigma = 0.05$.

\paragraph{Scale}
Random scaling is applied independently to the X, Y, and Z axes, with the scaling coefficient for each axis sampled as $s \sim \mathcal{U}(1/S, S)$. Here, $S \in \{1.6, 1.7, 1.8, 1.9, 2.0\}$ defines the five levels of severity. We chose the most severe level, $S = 2.0$, for our study.

\paragraph{Rotate}
Random rotations are applied using X-Y-Z Euler angles $(\alpha, \beta, \gamma)$, where $\alpha, \beta, \gamma \sim \mathcal{U}(-\theta, \theta)$ and $\theta \in \{\pi/30, \pi/15, \pi/10, \pi/7.5, \pi/6\}$ represent the five severity levels. We selected the most extreme rotation range, $\theta = \pi/6$.

\paragraph{Drop Global}
Points in the cloud are randomly shuffled, and the last \( N \cdot \rho \) points are dropped, where \( N = 1024 \) represents the total number of points, and \( \rho \in \{0.25, 0.375, 0.5, 0.675, 0.75\} \) defines the severity levels. We used only the highest severity level, $\rho = 0.75$.

\paragraph{Drop Local}
A total of $K$ points are removed, where $K \in \{100, 200, 300, 400, 500\}$ represents the severity levels. Clusters of points are identified by randomly selecting $C \sim \mathcal{U}\{1, 8\}$ local centers. For each cluster, $N_i$ points are removed such that $K = \sum_{i=1}^C N_i$. We used the most severe level, $K = 500$.

\paragraph{Add Global}
A total of \( K \) points were uniformly sampled from a unit sphere and added to the point cloud, where \( K \in \{10, 20, 30, 40, 50\} \) defines the severity levels. Only the highest level, $K = 50$, was included in our experiments.

\paragraph{Add Local}
A total of \( K \) points were added, with \( K \in \{100, 200, 300, 400, 500\} \) representing the severity levels. Randomly shuffled points were used to generate $C \sim \mathcal{U}\{1, 8\}$ local clusters, and $N_i$ points were added to each cluster such that $K = \sum_{i=1}^C N_i$. The coordinates of these points were generated from a Normal distribution $\mathcal{N}(\mu_i, \sigma_i^2)$, where $\mu_i$ is the cluster center's coordinate and $\sigma_i \in \{0.075, 0.1, 0.125\}$. We only applied the highest severity level, $K = 500$.

\paragraph{Post-processing}
Point cloud classification models require the number of points in point clouds within a batch to be consistent for batch training and inference. The corruptions of scale, jitter, and rotate do not alter the number of points in the original point cloud (1024), meaning the corrupted point clouds used for testing and training directly consist of artificially corrupted point clouds. However, for Add Global and Add Local corruptions, the number of points in the corrupted point clouds increases. Conversely, for Dropout Global and Dropout Local, the number of points in the corrupted point clouds decreases. 

Point clouds with a number of points other than 1024 cannot be directly concatenated into batched data with point clouds containing 1024 points. Therefore, these four types of corrupted point clouds cannot be used directly for training. Instead, the number of points is fixed at 1024. For point clouds with more than 1024 points, farthest point sampling is applied to remove the excess points. For point clouds with fewer than 1024 points, one random point is duplicated to ensure a fixed number of 1024 points. This approach is designed to minimize the impact of duplicate points on k-NN local aggregation. 

Through this process, the corrupted point clouds are adjusted to have a fixed number of points. This approach is reasonable, as many point cloud classification networks process point clouds by fixing the number of points or the dimensions of k-NN local aggregation data to 1024 through farthest point sampling.

\section{Farthest point sampling}
Processing point clouds often requires transforming the data into a uniform format, such as ensuring each sample contains the same number of points, to suit downstream tasks like classification and segmentation. A naive approach is to randomly select a fixed number of points. However, this method tends to lose critical shape information and fails to capture the global structure of the object, making it an inefficient way to generate representative data for model inputs.

To address these limitations, Farthest Point Sampling (FPS) was proposed. FPS iteratively selects the farthest point from previously chosen points, effectively creating a subset of points that represent the overall structure of the object. This approach ensures that the selected points are evenly distributed across the object's surface, capturing the complete geometry with a limited number of points. Compared to random sampling, FPS preserves the geometric and structural features of the object in a simple manner.

The computational complexity of FPS is \(\mathit{O}(m \cdot n)\), where \(n\) is the number of points in the original point cloud and \(m\) is the number of sampled points. Despite its relatively high computational cost, FPS is widely used in 3D point cloud applications due to its ability to generate high-quality samples for tasks like object classification, segmentation, and reconstruction.

The specific steps of FPS are detailed in Algorithm~\ref{fps}:

\begin{algorithm}
\caption{Farthest Point Sampling}
\label{fps}
\begin{algorithmic}[1]
\Require Point cloud \( \mathcal{P} = \{p_1, p_2, \dots, p_N\} \), number of samples \( M \)
\Ensure Sampled points \( \mathcal{S} \) with \( |\mathcal{S}| = M \)

\State Initialize sampled set \( \mathcal{S} = \{\} \)
\State Initialize distance array \( \text{distances} = \infty \cdot \mathbf{1}_{N} \)
\State Randomly select an initial point \( p_0 \) from \( \mathcal{P} \)
\State Add \( p_0 \) to \( \mathcal{S} \)

\While{\( |\mathcal{S}| < M \)}
    \ForAll{\( p_i \in \mathcal{P} \setminus \mathcal{S} \)}
        \State \( d_i = \|p_i - p_{\text{last}}\|^2 \) 
        \State \( \text{distances}[i] \leftarrow \min(\text{distances}[i], d_i) \) 
    \EndFor

    \State Select \( p_{\text{next}} = \arg\max_{p_i \in \mathcal{P} \setminus \mathcal{S}} \text{distances}[i] \) 
    \State Add \( p_{\text{next}} \) to \( \mathcal{S} \)
\EndWhile

\Return \( \mathcal{S} \)
\end{algorithmic}
\end{algorithm}

As seen in Algorithm~\ref{fps}, FPS ensures that the farthest point from the current sampled set is selected at each iteration. While effective, this approach can sometimes select outliers, as they often satisfy the selection criteria but do not contribute meaningful information for downstream tasks. To mitigate this issue, instead of always selecting the farthest point, our method assigns higher selection probabilities to points farther away from the current point and lower probabilities to closer points. This strategy reduces the likelihood of selecting outliers while still maintaining a diverse and representative set of sampled points. This approach balances robustness and diversity, ensuring better performance in downstream tasks.

\section{Ablation studies}
In this section, we conducted several ablation studies. First, we investigated the impact of the two modules in our method separately, Farthest Exemplar Selection (FES) and Replay with Downsampling, on replay-based methods. Then, we explored whether the improvement brought by Replay with Downsampling remains stable under different replay buffer memory sizes \( M \).

\paragraph{Ablation on Modules}
We tested the individual effects of FES and Replay with Downsampling on two baselines and two datasets, respectively. As shown in Table \ref{tab3}, we observed that both modules significantly improve the performance of the baselines when used independently, with FES generally providing a greater improvement. The hyperparameters used here are consistent with those in the main experiments, so the values of \( \alpha \) and \( r \) are not optimized. Similar to the main experiments, we found that both modules lead to greater improvements in CIL tasks with a larger number of tasks. This indicates that both modules enhance the effectiveness of replay samples.

\begin{table*}[!ht]
\centering
\scalebox{1.0}{
\begin{tabular}{llcccccc}
\toprule
\multicolumn{1}{c}{\multirow{2}{*}{Metric}} & \multicolumn{1}{c}{\multirow{2}{*}{Method}} & \multicolumn{3}{c}{ModelNet40-Mix}             & \multicolumn{3}{c}{OmniObject3D-Mix}           \\ \cmidrule(lr){3-5} \cmidrule(lr){6-8} 
\multicolumn{1}{c}{}                        & \multicolumn{1}{c}{}                       & 4-task   & 10-task  & 20-task  & 4-task    & 10-task   & 20-task \\ \midrule
\multirow{8}{*}{OA(\%) \(\uparrow\) }       & Replay                         & 65.05    & 59.99    & 56.63    & 44.68    & 42.79     & 41.97   \\
                                            & DER                            & 78.19    & 74.87   & 70.07    & 53.38     & 52.57    & 52.14   \\  
                                            \cmidrule(lr){2-8} 
                                            & Downsampling only + Replay     & 67.44    & 63.50    & 60.75    & 46.34     & 45.30     & 44.82    \\
                                            & Downsampling only + DER        & 80.17    & 77.92    & 73.48    & 54.26     & 53.47     & 53.31     \\
                                            \cmidrule(lr){2-8} 
                                            & FES only + Replay              & 70.74    & 65.23    & 63.34    & 47.66     & 46.60     & 46.24   \\
                                            & FES only + DER                 & 81.24    & 79.46    & 76.01    & 55.41     & 54.90     & 54.33   \\ 
                                            \cmidrule(lr){2-8} 
                                            & Ours + Replay                  & 72.82    & 69.31    & 67.78    & 49.59     & 48.18     & 48.03   \\
                                            & Ours + DER                     & 83.57    & 81.27    & 78.13    & 56.05     & 56.75     & 57.15   \\ 
                                            \bottomrule
\end{tabular}
}
\caption{Ablation studies on Farthest Exemplar Selection and Replay with Point Cloud Downsampling}
\label{tab3}
\end{table*}
\paragraph{Ablation on replay buffer memory size}

Here are the definitions of some important varaible in replay with point cloud downsampling.

\begin{itemize}
    \item \textbf{Downsampled Point Cloud Ratio} \( \alpha \): The fraction of downsampled point clouds within the replay buffer for each class, defined as:
    \[
    \alpha = \frac{|\{P_{t,i}^{\text{down}}\}_{i=1}^{n_t^{\text{rb}}}|}{|\mathcal{D}_{t}^{\text{rb}}|}
    \]
    where \( |\{P_{t,i}^{\text{down}}\}_{i=1}^{n_t^{\text{rb}}}| \) is the number of downsampled point clouds, and \( |\mathcal{D}_{t}^{\text{rb}}| \) is the total number of samples in the replay buffer.

    \item \textbf{Downsampling Rate} \( r \): The ratio of the number of points in the original point cloud to the number of points in the downsampled point cloud, defined as:
    \[
    r = \frac{|P_{t,i}|}{|P_{t,i}^{\text{down}}|}
    \]
    where \( |P_{t,i}| \) and \( |P_{t,i}^{\text{down}}| \) are the number of points in the original and downsampled point clouds, respectively.

    \item \textbf{Replay Buffer Memory Size} \( M \): The total number of points that can be stored in the replay buffer for each class.
\end{itemize}

In the main paper, we analyzed the trade-off between point cloud quality and quantity, focusing on the impact of \( \alpha \) and \( r \) on model performance. In this subsection, we varied the value of \( M \) to analyze the changes in model performance improvements brought by point cloud downsampling under different parameter combinations. All experiments in this subsection were based on the ModelNet-Mix dataset, used Replay + FES, and fixed the number of tasks at 10.

As shown in Table~\ref{tab4}, by observing each row horizontally, we found that increasing \( M \) generally led to improved model performance, which was consistent with intuition. When comparing different rows, we observed that when \( r \) was small, meaning the quality of the downsampled point clouds was higher, the performance improvement remained stable across different values of \( M \). However, when \( r \) was large, meaning the quality of the downsampled point clouds was lower, the performance improvement was more significant for larger \( M \), while the improvement was less noticeable for smaller \( M \).

\begin{figure*}[ht]
    \centering
    \includegraphics[width=\linewidth]{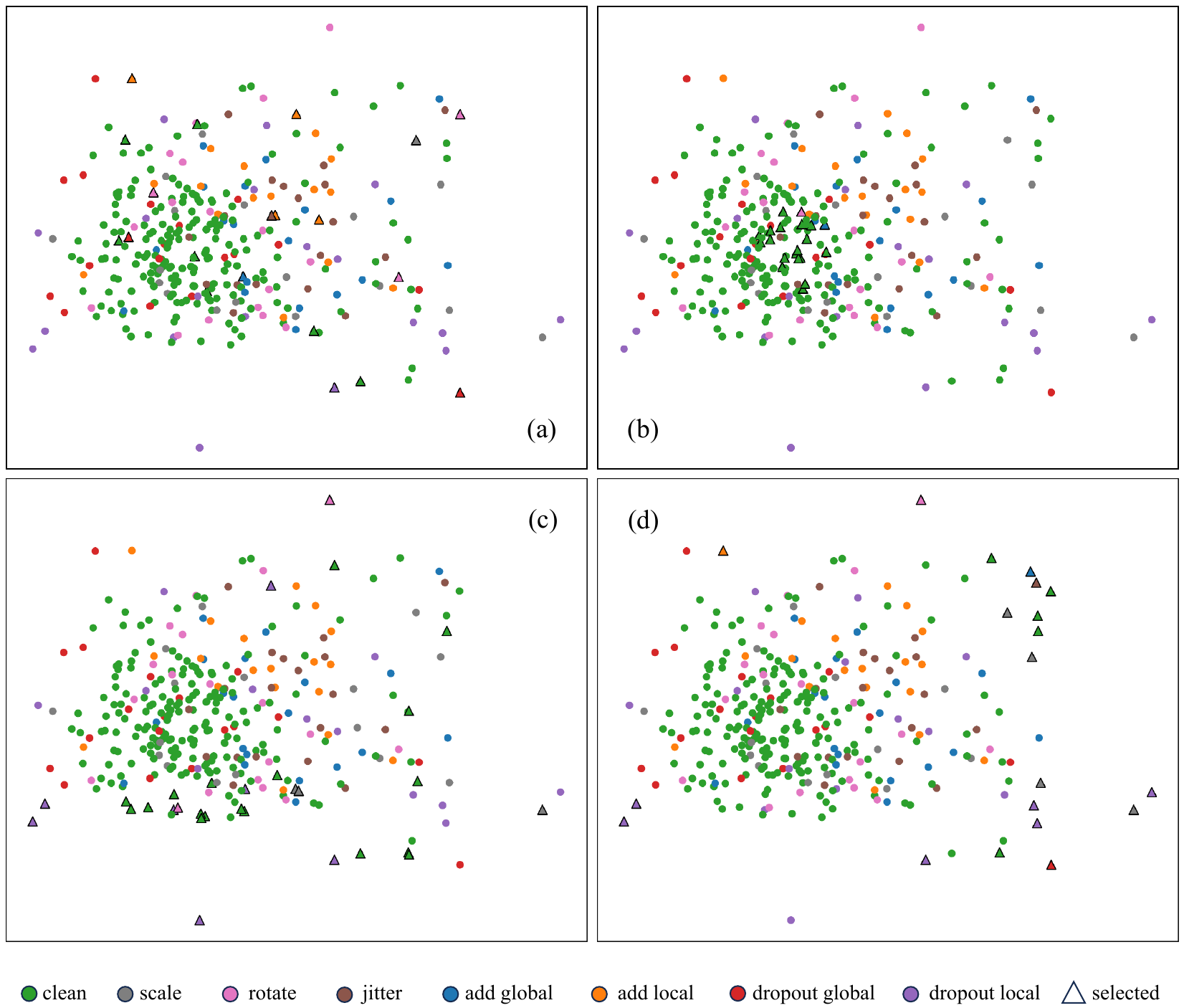}
    \caption{We visualized the performance of several non-learning exemplar selection methods in the point cloud feature space using PCA: (a) Farthest(ours), (b) Herding, (c) Entropy, and (d) Distance Exemplar Selection. Different colors are used to represent various corrupted point cloud features within a class (for visualization purposes only, unseen by the model). We observe that Herding focuses excessively on selecting central samples within the class, with most selected samples being clean point clouds. Meanwhile, Entropy and Distance focus on samples at the distribution's margin, overlooking densely distributed feature regions. Our farthest exemplar selection, on the other hand, better captures the intra-class diversity of this category in feature space.(Please zoom in for a better view)}
    \label{fig7}
\end{figure*}

\begin{table*}[t]
\centering
\scalebox{1.0}{
\begin{tabular}{lccccc}
\toprule
\multicolumn{1}{c}{Metric}                        & \multicolumn{1}{c}{Downsampling param.}          & M=2      & M=4      & M=10     & M=20 \\ \midrule
\multirow{4}{*}{OA(\%) \(\uparrow\) }       & $\alpha$=0, r=1       & 36.13    & 44.00    & 56.23    & 65.23 \\
                                            & $\alpha$=0.5, r=2               & 40.97    & 48.80   & 60.11    & 69.31 \\
                                            & $\alpha$=0.5, r=4               & 40.13    & 48.78    & 60.68    & 71.03 \\
                                            & $\alpha$=0.5, r=8               & 36.10    & 44.74    & 59.88    & 67.98 \\
                                            \bottomrule
\end{tabular}
}
\caption{Ablation studies on the replay buffer memory size }
\label{tab4}
\end{table*}

\section{Impact on Different Types of Corruptions}

\begin{table*}[t]
\centering
\scalebox{0.9}{
\begin{tabular}{lcccccccccc}
\toprule
\multicolumn{1}{c}{Metric}                  & \multicolumn{1}{c}{Method}      & overall   & clean    & S        & J        & R       & $A_G$    & $A_L$    & $D_G$  & $D_L$ \\ \midrule
\multirow{2}{*}{acc(\%) \(\uparrow\) }       & replay                         & 59.99     & 83.01    & 67.88    & 54.33    & 51.59   & 76.53    & 62.39    & 44.75  & 39.44 \\
                                             & Ours+replay                    & 69.31     & 88.98    & 71.10    & 62.58    & 59.17   & 79.75    & 73.84    & 67.98  & 51.11 \\
                                            \bottomrule
\end{tabular}
}
\caption{Impact of Our Method on Different Types of Corruptions. 
The table shows the accuracy (\%) of the baseline replay method and our proposed method with replay on various types of corruptions. Notations: S: Scale; J: Jitter; R: Rotate; $A_G$: Add Global;$A_L$: Add Local; $D_G$: Dropout Global $D_L$: Dropout Local}

\label{tab5}
\end{table*}
For simplicity and in line with the tradition of CIL research, we chose accuracy on the test set as the sole evaluation metric in the experiments of the main text. In this subsection, we aimed to demonstrate the specific improvements of our method over the baseline model on point clouds with different types of corruptions. As shown in the Table \ref{tab5}, we observed that the baseline model performed well on clean point clouds but poorly on corrupted point clouds. This could be due to two reasons: (1) corrupted point clouds were inherently more difficult to classify, and (2) corrupted point clouds suffered more from catastrophic forgetting.

In the main text, we visualized the exemplar selection process of the baseline model and found that corrupted point clouds, being farther from the class feature center, were less likely to be selected. Our method not only improved the performance on corrupted point clouds but also enhanced the performance on clean point clouds. This indicated that the diversity within clean point clouds was also an important factor affecting robustness. By selecting replay samples that reflected in-class diversity and using point cloud compression to store more samples, our method improved robustness to various corrupted point clouds under the Robust3D-CIL setting.

\clearpage

\end{document}